\documentclass[10pt,twocolumn,letterpaper]{article}
\usepackage[pagenumbers]{cvpr} %

\usepackage{cuted} %

\usepackage{currfile} %

\usepackage{caption} %

\providecommand{\supp}[2]{#1}

\usepackage{booktabs}
\usepackage{array}
\usepackage{amsmath,amssymb}
\definecolor{cvprblue}{rgb}{0.21,0.49,0.74}
\usepackage[pagebackref,breaklinks,colorlinks,allcolors=cvprblue]{hyperref}
\def\confName{arXiv}
\def\confYear{2026}
\title{Right Look, Wrong Destination: What Velocity-Field Models of Perturbation Cannot See Yet}
\author{Abhishek Sharma$^{1}$ \qquad Aleksej Zelezniak$^{1,2,3}$\\[4pt]
{\small $^{1}$Randall Centre for Cell \& Molecular Biophysics, King's College London, London, UK}\\
{\small $^{2}$Department of Life Sciences, Chalmers University of Technology, Gothenburg, Sweden}\\
{\small $^{3}$Institute of Biotechnology, Life Sciences Centre, Vilnius University, Vilnius, Lithuania}\\[2pt]
{\small \href{mailto:abhishek.1.sharma@kcl.ac.uk}{\tt abhishek.1.sharma@kcl.ac.uk}, \href{mailto:aleksej.zelezniak@kcl.ac.uk}{\tt aleksej.zelezniak@kcl.ac.uk}}}
\renewcommand{\supp}[2]{#2}
\begin{document}
\maketitle
\begin{strip}
\centering
\includegraphics[width=\linewidth]{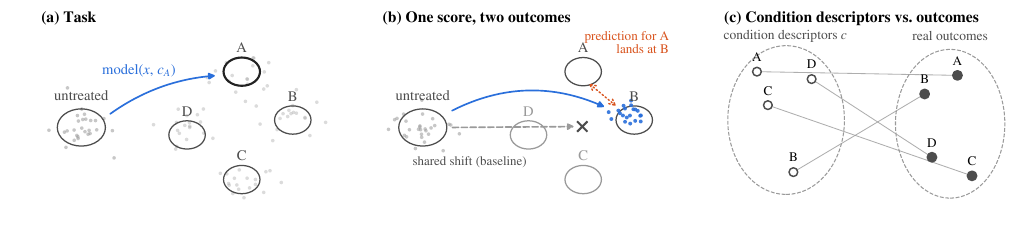}
\captionsetup{hypcap=false}
\captionof{figure}{\textbf{Predicting an unseen condition.} \textbf{(a)} A conditional generative model predicts the cell state under an unseen condition $A$ from untreated cells and its descriptor $c_A$. \textbf{(b)} A set-level score can be equally high when a prediction conditioned on $A$ instead resembles another condition $B$. A condition-blind baseline, based on the response shared across conditions, further illustrates how much predictive performance can be achieved without condition-specific information. \textbf{(c)} Similarity in condition-descriptor space does not necessarily translate into similarity of cellular outcomes.}
\captionsetup{hypcap=true}
\label{fig:story}
\end{strip}

\begin{abstract}
Conditional generative models that learn velocity fields over cell states to predict intervention effects have recently emerged across several areas of biology, including gene perturbation prediction, drug-treated cell imaging, protein dynamics, and cell reprogramming. However, their evaluations lack strong, interpretable baselines, which makes it difficult to establish what these models learn and where progress is needed. In gene-perturbation prediction, mean and linear models have become essential benchmarks for evaluating new methods. We extend this principle to conditional generative models, introducing interpretable baselines and three diagnostics. Our results show that simple baselines compete with or exceed published models on several of their own benchmarks. Moreover, although some models improve over condition-blind baselines, their predictions for unseen conditions are often closer to the outcomes of other conditions than to the intended ones. A model-free analysis reveals a potential common limitation: distances between condition descriptors are uncorrelated with distances between their corresponding observed outcomes.
\end{abstract}

\section{Introduction}
\label{sec:intro}

Predicting cellular responses to previously untested interventions is a central problem in computational biology~\cite{waddington1957,schiebinger2019wot}. Given measurements of untreated cells and a description of a new drug, gene deletion or combination of reprogramming factors~\cite{takahashi2006ipsc,vierbuchen2010}, the goal is to generate the resulting cellular states. Solving this problem could help prioritise experiments before they are performed.

Conditional generative models offer a promising approach: they learn a velocity field over cell states and integrate it to predict the state that an intervention induces. Three of the models we study learn this field from untreated to treated cells; the fourth, NAVIGO, learns it from normal development and imposes the intervention only when the field is integrated. Recent applications include CellFlow~\cite{klein2025cellflow} for gene-perturbation prediction, CellFlux~\cite{zhang2025cellflux} for drug-treated cell imaging, ProteinTalks~\cite{proteintalks2026} for protein dynamics following drug treatment, and NAVIGO~\cite{navigo2026} for cell-fate prediction and reprogramming. These approaches use flow matching~\cite{lipman2023flow,tong2024cfm} or neural ordinary differential equations (ODEs)~\cite{chen2018neuralode} to learn and integrate the resulting dynamics.

\begin{table}[t]
\centering\small
\begin{tabular}{@{}l>{\raggedright\arraybackslash}p{0.58\linewidth}@{}}
\toprule
Biology & Definition in this paper \\
\midrule
\textbf{Gene expression} & The activity of each gene in a cell, measured as a count; represented as a vector with approximately 20{,}000 entries. \\
\textbf{Protein} & A molecule produced by a gene that performs cellular functions; protein levels form a vector with one entry per protein. \\
\textbf{Cell} & One sample, represented by a gene-expression vector, protein-level vector or image. \\
\textbf{Control cells} & Cells receiving no intervention; the source distribution. \\
\textbf{Condition} & An intervention applied to a group of cells, such as a drug, gene deletion or set of forced genes (reprogramming factors); the condition label. \\
\textbf{Condition descriptor} $c$ & A fixed, externally supplied representation of a condition, such as a protein-sequence embedding or chemical fingerprint. \\
\textbf{Held-out condition} & A condition whose cells are excluded from training; at test time, the model receives only its descriptor $c$. \\
\textbf{Reprogramming} & Forcing the expression of selected factors to convert a cell into another cell type. \\
\bottomrule
\end{tabular}
\caption{\textbf{Glossary of biological terms used in this paper.}}
\label{tab:glossary}
\end{table}

Evaluation typically relies on set-level distribution metrics, such as energy distance and FID, or AUROC on held-out conditions, with comparisons focused primarily on other learned models or existing trajectory and network-inference methods. Simple baselines are reported unevenly. This leaves an important question unanswered: does a model actually predict the response specific to a new intervention, beyond the response shared across conditions?

The lack of strong, interpretable baselines hinders progress: a single benchmark score cannot reveal whether a model has learned the response shared across treatments, the response specific to a new condition, or neither (Fig.~\ref{fig:story}).

Gene-perturbation prediction illustrates the value of such baselines. Ahlmann-Eltze \etal~\cite{ahlmannEltze2025linear} showed that deep models did not outperform simple mean and linear baselines, a finding reinforced by broader benchmarks~\cite{csendes2025benchmarking,wenteler2024perteval}. Reporting these baselines is now an expected reference for new perturbation-prediction methods. Conditional generative models, however, lack a comparable standard.

We introduce interpretable baselines and three diagnostics for evaluating conditional generative models of perturbation, and apply them to four published models using their original datasets, released checkpoints where available, and evaluation metrics. Our baselines compete with or exceed the published models on several of their own benchmarks. For NAVIGO~\cite{navigo2026}, a linear model matches or exceeds the published score on the factor-set benchmark. CellFlow~\cite{klein2025cellflow} improves over condition-blind baselines, yet for both CellFlow and CellFlux~\cite{zhang2025cellflux} the predictions for unseen conditions land nearest the intended outcome in at most a quarter of cases. We also propose a model-free analysis that points to a common limitation: condition-descriptor distances are uncorrelated with distances between the corresponding observed outcomes. Across these models, capturing condition-dependent structure does not necessarily translate into generalisation to unseen condition-specific outcomes.

Our contributions are threefold:
\begin{enumerate}
\item \textbf{An evaluation protocol for conditional generative models.} We propose interpretable baselines and three complementary diagnostics, reported alongside the model's original metric. Together, they show whether a model predicts only the response common to all treatments, the outcome of the intended condition, or the outcome of a different, similar condition (Section~\ref{sec:protocol}).
\item \textbf{Effective baselines.} Using the original benchmarks of four published models, we show that simple baselines compete with or exceed the published models in several cases, while predictions for unseen conditions frequently favour a different condition (Section~\ref{sec:results}).
\item \textbf{A model-free analysis of condition descriptors.} We show that descriptor similarity is uncorrelated with outcome similarity, suggesting that limited information in the input contributes to the generalisation gap, beyond limitations of the generator alone (Section~\ref{sec:why}).
\end{enumerate}

\section{Related Work}
\label{sec:related}

\paragraph{Conditional generative models of perturbation.}
CellFlow~\cite{klein2025cellflow} uses conditional flow matching to predict single-cell gene-activity profiles following genetic perturbations. CellFlux~\cite{zhang2025cellflux} applies flow matching to $96{\times}96$ microscopy images of individual drug-treated cells, conditioning generation on chemical fingerprints and control images. ProteinTalks~\cite{proteintalks2026} models protein-level dynamics following drug treatment using neural ODEs. NAVIGO~\cite{navigo2026} learns developmental cell-state dynamics through flow matching and applies them to predict reprogramming by combinations of transcription factors.

\paragraph{Baselines and evaluation of perturbation prediction.}
Simple baselines already rival deep models in gene-perturbation prediction~\cite{ahlmannEltze2025linear,csendes2025benchmarking,wenteler2024perteval}. Gijsen \etal~\cite{gijsen2026hallucinate} further showed that, for unseen pairs of perturbations, conditional flow models predict interaction effects (deviations from additive single-perturbation effects) that are largely unsupported by the observed data, a failure that aggregate error metrics do not reveal. We complement these studies by evaluating published generative models on their original benchmarks, focusing on generalisation to individual unseen conditions rather than aggregate endpoint accuracy. We introduce interpretable baselines and diagnostics that distinguish shared from condition-specific responses, and investigate whether the condition descriptors themselves capture similarities in observed outcomes.

\section{Background: Data and Model Formulation}
\label{sec:background}

\paragraph{Data.}
We represent each experimental condition as a set of cell measurements (Table~\ref{tab:glossary}). For condition $k$, let $Y_k=\{y_i\}_{i=1}^{n_k}$ denote the $n_k$ cells measured after the intervention, and let $X=\{x_j\}_{j=1}^{n_0}$ denote the $n_0$ matched control cells from the same experimental run, accounting for run-specific offsets. Every cell is a vector in the same space, $x_j, y_i \in \mathbb{R}^{d}$. Depending on the dataset, each cell is represented by a gene-activity vector (typically reduced by PCA to $d=50$--$100$), a protein-abundance vector ($d=5{,}585$ in ProteinTalks), or a microscopy image ($d = 96{\times}96{\times}3$). Since measurement destroys the cell, no individual cell can be observed both before and after intervention.

\paragraph{Task.}
Given a control cell $x \in \mathbb{R}^{d}$ and a condition descriptor $c_k \in \mathbb{R}^{m}$, a model $f:\mathbb{R}^{d}\times\mathbb{R}^{m}\to\mathbb{R}^{d}$ maps each control cell to a predicted treated cell; applied to all of $X$, it gives the predicted set $\hat{Y}_k=\{f(x_j,c_k)\}_{j=1}^{n_0}$. Conditions are partitioned into training and held-out sets. Each condition is represented by a fixed descriptor computed independently of the experimental outcomes: an ESM2 protein-sequence embedding for gene perturbations ($m=2{,}560$), a Morgan fingerprint ($m=1{,}024$ substructure counts) or a protein-binding profile with one entry per measured protein ($m=5{,}585$) for drugs. We distinguish throughout between the condition, which is the actual intervention, and its descriptor $c_k$, which is the only information about that intervention available to the model. Thus, generalisation to held-out conditions relies on their descriptors rather than observed responses.

\paragraph{Models.}
All four models share a common formulation: a network $v_\theta$ defines a velocity field over cell states, and predictions are obtained by integrating
\begin{equation}
\frac{dz}{dt} = v_\theta(z, t, c_k), \qquad z(0) = z_0(x),
\label{eq:ode}
\end{equation}
from an initial state $z_0(x)$, which is a control cell $x$ or an encoding of it, to a final time $T$; the prediction $f(x,c_k)$ is $z(T)$ or a readout of it. The models differ in how $v_\theta$ is trained. Flow matching~\cite{lipman2023flow,tong2024cfm} (CellFlow, CellFlux, NAVIGO) regresses $v_\theta$ against the target velocity along a prescribed interpolation path between paired source and target samples, without solving the differential equation during training. By contrast, a neural ODE~\cite{chen2018neuralode} (ProteinTalks) integrates Eq.~\eqref{eq:ode} during training and fits the resulting trajectory to measurements at observed time points. In NAVIGO, the velocity field is learned from normal development and takes no descriptor, while the condition is introduced by holding the selected factors fixed in $z$ during integration. Our baselines and diagnostics operate only on the inputs and outputs of $f$, making them applicable to all four models.

\section{Interpretable Baselines and Diagnostics}
\label{sec:protocol}

We introduce baselines that isolate different sources of predictive performance:
\begin{itemize}
\item \textbf{B1, untreated cells} (time-matched for time courses): how much of the prediction can be explained without modelling the intervention, or by time alone?
\item \textbf{B2, shared treatment response:} adding the mean shift across other conditions, or pooling their treated cells, measures how much performance comes from effects shared across interventions.
\item \textbf{B3, seen-condition samples:} how well does the model perform when predicting conditions already represented in training?
\item \textbf{B4, linear conditional model:} how much can be achieved by directly predicting the expected outcome from the condition descriptor, without a generative component?
\end{itemize}
We evaluate each model against the baselines applicable to its task (Section~\ref{sec:results}), including several baselines proposed in the original papers.

\paragraph{Notation.}
Let $A$ denote a held-out condition, with real samples $Y_A$ and model predictions $\hat{Y}_A$. Let $\mathcal{B}$ denote alternative conditions measured in the same context (\eg, the same time point, plate, or batch). We use the energy distance $E(\cdot,\cdot)$~\cite{szekely2013energy} to compare sample distributions, computed in the evaluation space used by each original paper: the model's latent cell space for CellFlow and InceptionV3 pooled features (the feature space underlying FID) for CellFlux.

\paragraph{Diagnostic 1 (D1): Positive control (are conditions distinguishable in the data?).}
We randomly split the real samples of condition $A$ into two halves, $Y_A^{(1)}$ and $Y_A^{(2)}$. A condition is considered distinguishable only if
\begin{equation}
E(Y_A^{(1)},Y_A^{(2)}) < E(Y_A^{(1)},Y_B), \quad \forall B\in\mathcal{B}.
\label{eq:fair}
\end{equation}
This establishes whether the experimental data contain sufficient information to distinguish $A$ from alternative conditions. If even independent samples from the same condition cannot be distinguished from those of another, a model cannot reasonably be expected to separate them. The criterion is deliberately stringent, since $A$ must be distinguishable from every alternative. D2 and D3 are evaluated only on the conditions that pass it.

\paragraph{Diagnostic 2 (D2): Wrong destination (does the prediction match the intended condition?).}
We rank the real condition distributions by their distance to the model's prediction:
\begin{equation}
r_A = 1 + \bigl|\{B \in \mathcal{B} : E(\hat{Y}_A, Y_B) < E(\hat{Y}_A, Y_A)\}\bigr|.
\label{eq:rank}
\end{equation}
A rank of $r_A=1$ means that the prediction is closer to the intended condition than to any alternative. We consider a model to generalise successfully if it achieves $r_A=1$ for at least two-thirds of distinguishable held-out conditions. The threshold is a methodological choice: it asks for a clear majority and lies far above the $1/(|\mathcal{B}|+1)$ expected from a random ordering. We report the counts themselves, so that a reader can apply a different threshold.

\paragraph{Diagnostic 3 (D3): Condition swap (does the model use the condition descriptor?).}
We keep the input cells and random noise fixed while replacing the intended condition descriptor $c_A$ with that of another condition $c_B$. Let $\hat{Y}_{A\rightarrow B}$ denote the resulting prediction. We measure the effect of this intervention on the prediction:
\begin{equation}
s_{AB} = \frac{E(\hat{Y}_A, \hat{Y}_{A \to B})}{E(Y_A, Y_B)}.
\label{eq:swap}
\end{equation}
This measures whether swapping the condition descriptor meaningfully changes the predicted cell distribution, relative to the separation between the corresponding real conditions. A score of $s\approx 0$ indicates that the prediction is insensitive to the descriptor, whereas $s\approx 1$ indicates that the swap shifts the prediction by approximately the distance between the real conditions. We consider a model descriptor-sensitive if the median $s\geq 0.5$. Importantly, sensitivity does not imply correctness: a prediction may change substantially without moving towards the intended outcome. This is assessed separately by D2. Together, these diagnostics distinguish models that ignore condition descriptors from those that use them but predict the wrong outcome.

\paragraph{Descriptor check (do condition descriptors encode outcome information?).}
Across pairs of conditions, we compute the Spearman correlation $\rho$ between distances in descriptor space and distances between their real outcome distributions, $E(Y_A,Y_B)$. Statistical significance is assessed against a null distribution generated by 1{,}000 random permutations of condition labels. We consider descriptors informative if $\rho>0.3$ and $p<0.01$. This test is model-free: it determines whether the condition descriptors themselves capture differences in experimental outcomes, independently of any predictive model.

Baselines (B1--B4) are simple alternative predictors: copying the untreated cells, adding the mean response, or fitting a linear model. Each is evaluated using the paper's own metric, answering a straightforward question: does the published model outperform simple alternatives on its own benchmark? This is a direct comparison of predictive scores.

Diagnostics, by contrast, test what a prediction actually represents. Rather than producing predictions, they probe the model's outputs, the underlying data, or the descriptors used. Both evaluations are necessary: a model can outperform every baseline yet fail the diagnostics. For example, CellFlow beats B1 and B2, but its predictions identify the intended condition for only 1/9 conditions (Section~\ref{sec:results}). B3 connects the two: applying D2 to training conditions illustrates what successful diagnostic performance looks like.

\section{Experiments and Results}
\label{sec:results}

\begin{figure*}[t]
\centering
\includegraphics[width=\textwidth]{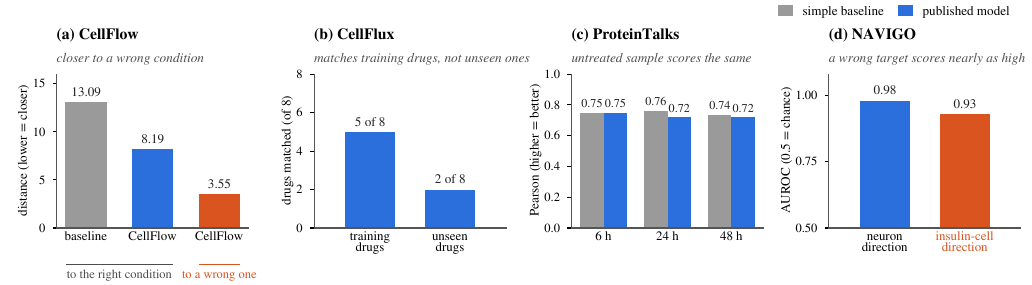}
\caption{\textbf{Published models against simple baselines}, each on its own data and metric. \textbf{(a)} CellFlow's predictions lie closer to the intended condition than the best condition-blind baseline (B2) does, but even closer to a wrong condition. \textbf{(b)} CellFlux matches 5/8 training drugs (B3) to their corresponding images, but only 2/8 unseen drugs (D2). \textbf{(c)} The untreated sample at the same time point (B1) scores as highly as ProteinTalks. \textbf{(d)} For NAVIGO's neuron factor sets, a different cell type's direction (0.93) approaches the trained model's score (0.98) along the neuron direction.}
\label{fig:results}
\end{figure*}

When comparing the published methods with simple baselines, we refer to the baselines of Section~\ref{sec:protocol} as B1--B4. Likewise, we refer to its three diagnostics as D1--D3. Before applying our tests, we reproduce each paper's headline metric using its original evaluation code\supp{ (Supp.\ Table~S1)}{}.

Two statistics recur below several times as a pair $(\rho, p)$. For the correlation $\rho$, a value near $0$ means that two quantities are unrelated, and a value near $1$ that they rise together. The $p$-value is the probability of obtaining a result at least this strong by chance alone: a small $p$ (below $0.01$) indicates a real effect, whereas a large $p$ means the result is consistent with chance.

\subsection{CellFlow: Gene Deletions in Whole Embryos}
\paragraph{Setup.} CellFlow predicts single-cell responses to gene deletions in zebrafish embryos~\cite{saunders2023zscape}. Each cell is represented by a gene-activity vector projected into the model's 100-dimensional input space. Conditions comprise one- or two-gene deletions measured at embryo ages of 18--72 hours, with age-matched unmodified embryos as controls. Each condition is described by the ESM2 embedding of the deleted genes' protein sequences. We hold out gene-deletion-by-age pairs, while retaining the same deletions at other ages and other deletions in training.

\paragraph{Retraining and results (Table~\ref{tab:main}, Fig.~\ref{fig:results}a).} As no trained checkpoints were released, we retrained the published architecture from scratch for each held-out condition using the paper's full training recipe (1M iterations). Across our nine held-out conditions, the mean error was 8.19, close to the reported 7.49 across 71 conditions\supp{ (Supp.\ Table~S1)}{}. The authors' shorter tutorial recipe yielded a similar error of 7.50 and the same qualitative conclusions.

CellFlow outperforms three condition-blind baselines---control cells (B1), controls shifted by the mean response across other conditions, and pooled cells from other conditions (both B2)---achieving an average error of 8.19 versus 13.1--16.8, and outperforming them individually in six of nine conditions. However, its predictions identify the correct held-out condition in only one of nine cases (D2); the median rank of the intended condition is 3. As a separate statistic, the distance averaged over the nine conditions is substantially smaller to the nearest incorrect condition (3.55) than to the intended condition (8.19). The nearest alternative is often biologically related: for example, a two-gene deletion is predicted to resemble deletion of one of its constituent genes. Descriptor swapping (D3) changes the predictions by 41--68\% of the corresponding real-condition distance (median for each of the three conditions tested, tutorial-recipe models), indicating that CellFlow does use the condition descriptor, even though it frequently predicts the wrong outcome.

\paragraph{Second dataset: Immune cells and signalling proteins.} CellFlow also predicts responses of human blood immune cells from 12 donors to 90 signalling proteins (cytokines)~\cite{parse2024pbmc}, again using ESM2 embeddings as condition descriptors. Following the authors' training recipe, we evaluated four held-out cytokines (IFN-$\beta$, IL-15, IL-13 and C5a) in every donor under two settings: $K{=}0$, where the cytokine is entirely unseen during training, and $K{=}1$, where it is observed in one other donor.

The two condition-blind baselines (B1, B2) reproduce the paper's reported results (\supp{supplement}{control cells give 29.47 against 29.58 reported, and the mean response across other cytokines gives 26.17 against 26.17}). The paper shows CellFlow's own score for this setting only as a bar in a figure, without a number, so we cannot check that our retrained model matches theirs. All 92 donor--cytokine conditions pass the positive control (D1). Nevertheless, CellFlow's predictions rank closest to the correct cytokine (D2) in none of the 48 $K{=}0$ or 44 $K{=}1$ cases, with median ranks of 80 and 64, respectively, out of 90. More strikingly, predictions are closer to untreated cells than to the intended cytokine response in 46/48 and 33/44 cases. Descriptor swapping (D3) changes the output by only 11\% ($K{=}0$) and 21\% ($K{=}1$) of the corresponding real-condition distance, suggesting that the model largely ignores the condition descriptor and predicts little response. IFN-$\beta$ at $K{=}1$ is the only partial success (median rank 5). Consistent with these findings, the ESM2 embeddings carry little detectable information about cytokine outcomes ($\rho=-0.10$, $p=0.92$, across 4{,}005 cytokine pairs).

\paragraph{Summary.} The two datasets show different behaviours. On the gene-deletion benchmark, CellFlow improves over condition-blind baselines and responds to descriptor swaps, yet usually predicts a distribution closer to another, often related, deletion than to the intended one: it uses the descriptor without reliably reaching the intended outcome. On the immune-cell benchmark, it largely ignores the descriptor: predictions remain closer to untreated cells than to the target response in most cases, with little sensitivity to descriptor swaps. Thus, performance gains over condition-blind baselines do not necessarily indicate successful generalisation to unseen conditions.

\begin{table}[t]
\centering
\resizebox{\linewidth}{!}{%
\begin{tabular}{@{}lcccc@{}}
\toprule
 & D1: positive & D2: own cond. & D3: swap & Descr. \\
 & control & nearest & ratio & check \\
 & (data) $\uparrow$ & (model) $\uparrow$ & $s$ ($\to 1$) & $\rho$ $\uparrow$ \\
\midrule
CellFlow, zebrafish & 9/9 & \textbf{1/9} & 0.41--0.68 & 0.05 \\
CellFlow, immune, unseen & 48/48 & \textbf{0/48} & 0.11 & $-0.10$ \\
CellFlow, immune, 1 donor & 44/44 & \textbf{0/44} & 0.21 & $-0.10$ \\
CellFlux, held-out & 8/8 & \textbf{2/8} & 0.88 & $-0.04$ \\
CellFlux, training (B3) & 7/8 & 5/8 & 1.00 & -- \\
\bottomrule
\end{tabular}}
\caption{\textbf{Diagnostics on held-out conditions.} Training drugs (B3) shown for comparison; counts are conditions passing, out of all.}
\label{tab:main}
\end{table}

\subsection{CellFlux: Drug-Treated Cell Images}
\paragraph{Setup.} BBBC021~\cite{ljosa2012bbbc} contains single-cell microscopy images of human breast-cancer cells treated with 34 drugs or solvent-only controls. Each drug is represented by a 1{,}024-dimensional chemical fingerprint (Morgan substructure counts). CellFlux generates treated-cell images by integrating a fingerprint-conditioned flow from Gaussian noise and a control image from the same plate. The released checkpoint holds out eight drugs during training, as verified from its stored training arguments.

\paragraph{Results (Table~\ref{tab:main}, Fig.~\ref{fig:results}b).} To assess drug-specific generation, we select up to 224 test images for each of eight held-out and eight training drugs, generating each under all 16 drug descriptors while keeping the control image and noise fixed. Although all eight held-out drugs are distinguishable in the real data (D1), the correct drug descriptor yields the nearest matching outcome (D2) for only 2/8 drugs, compared with 5/8 for training drugs (B3; counted over all eight; nocodazole fails the positive control). Two held-out drugs rank 13th and 12th out of 16. Moreover, epothilone B is the nearest incorrect condition for 7/16 drugs.

\paragraph{Summary.} CellFlux distinguishes training drugs more reliably than unseen drugs: the correct outcome is nearest for 5/8 training drugs but only 2/8 held-out drugs. CellFlux nevertheless responds to unseen chemical descriptors (D3, swap ratio 0.88), suggesting that the limitation is not a failure to condition on the input, but a failure to generalise the mapping from chemical structure to drug-specific cellular phenotype.

\subsection{ProteinTalks: Drug Response over Time}
\paragraph{Setup.} ProteinTalks models drug responses in 18 breast-cancer cell lines using measurements of 5{,}585 proteins following treatment with 123 drugs or drug combinations, collected at 0, 6, 24 and 48 hours. A neural ODE predicts protein trajectories over time. Drug information enters only through the proteins known to bind each drug (at most 12; none for 397 of the 1{,}530 released experiments, whose predicted drug effect is therefore exactly zero), while a separate prediction head estimates drug efficacy, measured by its ability to inhibit cell growth.

\begin{table*}[t]
\centering
\small
\setlength{\tabcolsep}{11pt}
\begin{tabular}{@{}llll@{}}
\toprule
 & ProteinTalks & Reference & Answer \\
\midrule
\multicolumn{3}{@{}l}{\emph{1. Does the prediction change with the drug?}} & \textbf{Barely} \\
\quad spread across drugs, non-target proteins & $4{\times}10^{-9}$ & 1 (real data) & \\
\quad spread across drugs, target proteins & 0.02 & 1 (real data) & \\
\addlinespace[5pt]
\multicolumn{3}{@{}l}{\emph{2. Does it match its own drug?} (correlation, 48\,h)} & \textbf{No} \\
\quad with its own drug's response & $-0.03$ & & \\
\quad with other drugs' responses & $0.00$ & & \\
\addlinespace[5pt]
\multicolumn{3}{@{}l}{\emph{3. Does it beat the untreated sample (B1)?} (Pearson)} & \textbf{No} \\
\quad 6 / 24 / 48\,h & .750 / .720 / .718 & \textbf{.750 / .762 / .736} (untreated sample) & \\
\addlinespace[5pt]
\multicolumn{3}{@{}l}{\emph{4. What drives efficacy?} (AUROC)} & \textbf{The fingerprint} \\
\quad full model & 0.916 & & \\
\quad without predicted trajectories & 0.916 & & \\
\quad with shuffled fingerprints & 0.516 & 0.50 (chance) & \\
\bottomrule
\end{tabular}
\caption{\textbf{ProteinTalks on its 153 rebuilt test experiments.} Spread: variation of the predictions across drugs, as a fraction of the measured variation.}
\label{tab:proteintalks}
\end{table*}

\paragraph{Reproduction and results.} We reconstructed the paper's held-out test set using its released code, which randomly partitions the 1{,}530 experiments into 70/20/10 splits with seed 1995, and reproduced its preprocessing pipeline. All results below use the resulting 153 test experiments. The released checkpoint reproduces the reported drug-efficacy performance (\supp{Supp.\ Table~S1}{AUROC 0.916 against 0.92 reported}).

The predicted protein trajectories contain almost no drug-specific variation outside known drug targets. All drugs within a cell line start from the same untreated sample, and each protein's trajectory depends only on its initial level and whether it is a drug target. Consequently, non-target proteins can respond to drug identity only through normalisation statistics shared across proteins. Their predicted variation across drugs is a negligible fraction of the measured variation (Table~\ref{tab:proteintalks}, block~1), and the predicted drug-induced change is only 0.15\% of its measured magnitude. This near-zero variation is structural rather than a numerical precision artefact.

\paragraph{Drug-specific agreement (Table~\ref{tab:proteintalks}).} We next ask whether the predicted trajectories capture the correct drug-specific changes. For each experiment, we subtract the corresponding untreated trajectory from both measured and predicted trajectories, then remove the mean response across drugs within the same cell line to isolate drug-specific effects. We compare the resulting profiles over the 100 proteins with the largest measured drug-specific changes, using Spearman correlation. At 48\,h, each drug's prediction correlates no better with its own measured response than with those of other drugs in the same cell line (block~2; mean over up to 20 other drugs; 104 experiments). Restricting to the 69 test experiments whose drug has at least one known target gives the same result. Thus, predictions show no detectable agreement with their own drug-specific responses, despite the data being able to distinguish drug effects: the same drug in another cell line produces a more similar measured response than a different drug in that cell line (gap 0.086, $p=0.006$; at 24\,h, 0.077; not at 6\,h, 0.020).

\paragraph{Time-matched baseline and efficacy head (Table~\ref{tab:proteintalks}, Fig.~\ref{fig:results}c).} The model also fails to improve on the time-matched untreated baseline (B1). Taking the measured untreated sample from the same time point as the prediction, which ignores the drug entirely, ties or outperforms ProteinTalks on its own time-course metric (block~3\supp{; numbers in the supplement}{}). Moreover, the drug-efficacy head appears independent of the predicted protein trajectories: replacing them with drug-free trajectories leaves all three efficacy metrics unchanged. Conversely, shuffling drug fingerprints within each cell line drops efficacy prediction to chance (block~4), so it relies on the fingerprint rather than the predicted protein response. These findings are consistent across all 1{,}530 experiments.

\paragraph{Summary.} ProteinTalks reproduces its reported drug-efficacy performance, but its predicted protein trajectories add no detectable drug-specific information and do not outperform a time-matched untreated baseline (Table~\ref{tab:proteintalks}). The predicted variation is almost entirely restricted to known drug targets, while the efficacy head operates independently of the predicted trajectories and relies on drug fingerprints. Thus, the model predicts drug efficacy without learning the corresponding protein-level response dynamics.

\subsection{NAVIGO: Reprogramming One Cell Type into Another}
\paragraph{Setup.} NAVIGO~\cite{navigo2026} learns a flow over single-cell profiles from a developing mouse embryo. We evaluate its two released checkpoints: fibroblast-to-neuron and fibroblast-to-cardiomyocyte. Reprogramming is simulated by fixing selected transcription factors at three times their normal fibroblast expression and propagating cells through the flow. Predictions are scored by directional agreement with the fibroblast-to-target transition in the embryo atlas, measured over the 200 most upregulated and 200 most downregulated genes. We use NAVIGO's two published benchmarks: literature-derived factor sets for six target cell types and a laboratory screen of bHLH$\times$POU factor pairs for neuronal reprogramming~\cite{tsunemoto2018}.

\paragraph{Destination specificity (Table~\ref{tab:navigo}, Fig.~\ref{fig:results}d).} The directional metric does not establish that a model predicts the intended destination. We therefore score predictions against all six target directions in the atlas. Both checkpoints direct their standard cocktails most strongly towards their respective targets. The cardiomyocyte checkpoint also separates cardiac factor sets along the heart-muscle direction (0.75), substantially above all other directions (at most 0.52). For neurons, however, the factor-set benchmark is inconclusive: the trained checkpoint separates neuronal factor sets almost as well along the insulin-producing-cell direction (0.93) as along the neuronal direction (0.98).

\begin{table}[t]
\centering
\resizebox{\linewidth}{!}{%
\begin{tabular}{@{}llcc@{}}
\toprule
Checkpoint & Benchmark & Own direction & Best other direction \\
\midrule
Heart muscle & standard cocktail, score & \textbf{0.56} & 0.49 \\
Heart muscle & factor-set ranking, AUROC & \textbf{0.75} & 0.52 \\
Neuron & standard cocktail, score & \textbf{0.995} & 0.76 \\
Neuron & factor-set ranking, AUROC & 0.98 & 0.93 \\
Neuron & lab screen, 144 pairs, AUROC & 0.65 & 0.58 \\
\bottomrule
\end{tabular}}
\caption{\textbf{NAVIGO's released checkpoints on its own benchmarks.} Own target direction against the best of the five other directions.}
\label{tab:navigo}
\end{table}

\paragraph{Experimental benchmark.} The neuronal laboratory screen provides a more direct test against measured reprogramming outcomes. On NAVIGO's own evaluation set (144 factor pairs), the neuronal direction achieves an AUROC of 0.65, versus 0.58 for the best alternative direction (insulin-producing cells): a small gain of $+0.07$.\supp{}{ NAVIGO's published AUROC of 0.764 (42 headline cocktails, 6 of them positive) is reproduced only when the inputs are scaled as in the released inference code, which differs from the scaling used in training, and predictions are scored in the reverse direction (neurons to fibroblasts); the same predictions score 0.46 on all 144 scoreable cocktails.}

\paragraph{Linear baseline (B4).} Let $c \in \mathbb{R}^{F}$ hold the expression levels of $F$ transcription factors and $x$ the remaining genes. We fit a linear model of the conditional mean, $\hat{x}(c) = Wc + b$ (\supp{details in the supplement}{details in Appendix~\ref{sec:navigo_linear}}), and predict the effect of a cocktail as $W(c' - c)$, where $c'$ sets the cocktail's factors to a high level as in NAVIGO; the prediction is scored with NAVIGO's metric. On the six-target factor-set benchmark, one shared model fitted on the whole atlas achieves mean AUROC 0.82, increasing to 0.87 with one model per target fitted, following NAVIGO's design, on fibroblasts and that target cell type only, comparable to or above the approximately 0.76 that NAVIGO reports.

\paragraph{Summary.} NAVIGO learns target-associated developmental directions, most convincingly for cardiomyocytes, which is genuine evidence of target-associated structure. However, this structure does not clearly help to identify effective reprogramming factors: on the experimental screen the target direction gains only $+0.07$ AUROC over the best alternative, and a simple linear baseline matches or exceeds NAVIGO's reported factor-set performance. The distinction between directional alignment and successful reprogramming is central here: moving in the direction of a target is not the same as reaching it, and the published metric measures only the direction.

\section{What the Baselines Point To: Condition Descriptors}
\label{sec:why}

A model cannot reliably predict an unseen condition's outcome if its descriptor carries little information about that outcome. We test this independently of any model by asking whether similar descriptors correspond to similar observed effects.

For CellFlow, across 537 pairs of 71 gene-deletion conditions, ESM2 embedding distances are uncorrelated with distances between their observed outcomes ($\rho = 0.045$, shuffle $p = 0.26$). For CellFlux, across 120 pairs of 16 drugs, chemical fingerprint distances are likewise uncorrelated with distances between the corresponding real image sets ($\rho = -0.04$, $p = 0.61$). ProteinTalks provides a complementary case: drug-response predictions are determined by the chemical fingerprint alone, while its generated trajectories contain no drug-specific signal.

These results explain the patterns in Table~\ref{tab:main}. The models respond to condition descriptors, as demonstrated by the condition-swap experiments, but this responsiveness does not imply that the descriptors identify the corresponding outcomes. Instead, unseen conditions tend towards familiar training outcomes. For CellFlow, the prediction lands on the condition with the nearest descriptor for 3/9 held-out conditions; notably, the embedding's nearest neighbour also matches the nearest neighbour in outcome space for only 3/9. Thus, the limitation lies at least partly in the information available at the input: increasing model capacity or training duration alone is unlikely to resolve it. Our analysis shows only that outcome distance does not rise or fall steadily with descriptor distance. The descriptors may still hold information in a more complex, nonlinear form that the models fail to extract. Better condition descriptors that capture outcome-relevant structure are needed. Our descriptor checks use the descriptors the papers use. Other descriptors may carry more information; the descriptor check in Section~\ref{sec:protocol} tells a reader how to test them.

\section{Discussion and Conclusion}
\label{sec:discussion}

\paragraph{Where the published models still add value.}
The baselines do not explain all the structure captured by the published models. CellFlow is substantially closer to the true condition than any condition-blind baseline, outperforming them on 6/9 held-out conditions individually. NAVIGO's cardiomyocyte checkpoint separates cardiac factor sets along its intended developmental direction. CellFlux correctly ranks the outcome for 5/8 training drugs and responds strongly to changes in chemical fingerprints. These results show that learned generative models capture meaningful condition-dependent structure beyond simple baselines. However, capturing condition-dependent structure is not the same as generalising to unseen condition-specific outcomes.

\paragraph{An evaluation protocol.}
\label{sec:conclusion}
Gene-perturbation prediction now expects every new method to be compared with simple baselines~\cite{ahlmannEltze2025linear}. We propose that generative models of unseen conditions require similarly interpretable comparisons. We therefore propose an evaluation protocol, rather than a collection of additional metrics: alongside the primary benchmark metric, report simple baselines, a positive control, the rank of the correct condition among alternative conditions, and a condition-swap ratio. Together, they distinguish whether a model generates realistic outputs, responds to its condition descriptor, and actually predicts the intended condition-specific outcome. They require only an additional evaluation pass over predictions that models already generate.

Our findings highlight the need to evaluate not only whether generative models respond to condition descriptors, but whether they predict the specific outcomes of unseen interventions.

\section*{Data and code availability}
All datasets are public and are those used by the original papers. Notebooks reproducing every number in this paper are available at \url{https://github.com/bach-zouk/FlowsToWhere}.

\section*{Acknowledgements}
This work is supported by Biotechnology and Biological Sciences Research Council (BBSRC) grant number BB/Y000730/1.
Computational resources were provided by Isambard-AI, by King's Computational Research, Engineering and Technology
Environment (CREATE) at King's College London, and by the National Academic Infrastructure for Supercomputing in
Sweden (NAISS), funded by the Swedish Research Council.

\section*{Use of AI tools}
The research direction, methodology and ideas were set by the authors. Claude Code and ChatGPT were used to assist with scientific analysis, literature search, language editing, grammar and improving the clarity and conciseness of the manuscript. All scientific ideas, methodology, experiments, analyses, results, and conclusions are verified by the authors, who take full responsibility for the submitted content.

{\small
\bibliographystyle{ieeenat_fullname}
\bibliography{refs}
}

\appendix
\section{NAVIGO linear baseline}
\label{sec:navigo_linear}
The linear baseline (B4) of Section~4 predicts the expression of 2{,}634 genes from $F$ transcription factors (the factors of the benchmark cocktails and of the laboratory screen, plus the 100 factors ranked highest by a preliminary ridge fit; $F=137$ for the shared model and $F=157$ for the per-target models) and developmental time, $\hat{x}(c,t) = Wc + w_t t + b$. It is fitted by minimising the mean squared error, weighted by cell type as in NAVIGO's training data, with AdamW (learning rate $10^{-3}$, weight decay $10^{-4}$, batch size 4{,}096, 40 epochs). Each gene is min--max scaled. The model was fitted on our own rebuild of NAVIGO's mouse embryo atlas (NAVIGO's pipeline applied to all reads, 781{,}867 metacells) rather than on the released atlas file. The predicted effect of a cocktail is $W(c'-c)$ averaged over fibroblasts and is scored with NAVIGO's sign-agreement metric on each target's 200 most up- and downregulated genes. AUROCs are averaged over five bootstraps of 80\% of the fibroblasts and over the six targets. The shared model is fitted on all cells; each per-target model is fitted on fibroblasts and one target cell type and scored on that target only. NAVIGO's own score on this benchmark (approximately 0.76) is read from its Fig.~5b ($\pm 0.02$); it is not computed by us, because only two of its six per-target checkpoints are released.

\end{document}